# Can Language Models Analyze Data? Evaluating Large Language Models for Question Answering over Datasets

**Andreas Xenofontos[1], Pavlos Fafalios[1,2]**

[1]School of Production Engineering and Management, Technical University of Crete, Greece,

[2]Institute of Computer Science, Foundation for Research and Technology - Hellas, Greece.

***Abstract***

*This paper investigates the effectiveness of large language models (LLMs) in answering questions over datasets. We examine their performance in two scenarios: (a) directly answering questions given a dataset file as input, and (b) generating SQL queries to answer questions given the schema of a relational database. We also evaluate the impact of different prompting strategies on model performance. The study includes both state-of-the-art LLMs and smaller language models that require fewer resources and operate at lower computational and financial cost. Experiments are conducted on two datasets containing questions of varying difficulty. The results demonstrate the strong performance of large LLMs, while highlighting the limitations of smaller, more cost-efficient models. These findings contribute to a better understanding of how LLMs can be utilized in data analytics tasks and their associated limitations.*

## 1. Introduction

Large language models (LLMs) are language processing models trained on very large text corpora that are able to perform a wide range of tasks, such as text generation, summarization, translation, and question answering. Their recent development has attracted significant attention in both academia and industry, while surveys highlight their strong performance across many language understanding and generation tasks (Min et al., 2023). An important characteristic of LLMs is that they can interpret complex instructions expressed in natural language and produce both free text and structured outputs (such as code or queries).

Beyond traditional language processing tasks, there is increasing interest in examining whether LLMs can also support tasks that involve structured data analysis. In many research and real-world settings, answering questions over datasets requires specialised knowledge and technical skills. Researchers and analysts often need to understand database schemas, write SQL queries, or use programming environments such as Python to analyze data. As a result, non-technical users may face difficulties when attempting to explore datasets or retrieve specific information from them. Natural language interfaces have long been proposed as a way to simplify access to data by allowing users to express their information needs in natural language without requiring the use of formal query languages or coding.

The problem of answering natural language questions over structured data has been studied in several research areas, including semantic parsing, table question answering, and text-to-SQL. Early work focused on interpreting questions over semi-structured tables (Pasupat and Liang, 2015), while later research introduced neural approaches for generating SQL queries from natural language (Guo et al., 2019). Other work has explored models that can answer questions directly over tables using pre-trained architectures (Herzig et al., 2020). Despite significant progress, these tasks remain challenging, especially when questions require complex analysis or reasoning, or the combination of information from multiple sources.

Recent advances in LLMs have renewed interest in this area. Because of their ability to interpret instructions and generate structured outputs, they may serve as an intermediate layer between users and data sources. Recent studies have investigated the use of LLMs for tasks related to database querying and text-to-SQL generation (Gao et al., 2024), as well as data-related tasks such as data wrangling and transformation (Narayan et al., 2022). In parallel, research on prompting techniques has shown that carefully designed prompts can improve the reasoning capabilities of language models (Wei et al., 2022).

Two main approaches can be considered when using LLMs for question answering over datasets. In the first approach, the dataset itself is provided to the model, for example as a data file, and the model is asked to compute the answer to a question directly from the data. In the second approach, the model is given the schema of a relational database and asked to generate an SQL query that retrieves the correct answer. Although both approaches aim to bridge the gap between natural language questions and structured data, they rely on different capabilities of language models and may exhibit different limitations. In this paper, we investigate the ability of LLMs to answer questions over datasets under these two settings and examine the effect of different prompting strategies on model performance.

We examine both state-of-the-art large LLMs and smaller models that require fewer computational resources and may involve lower or no usage cost. This comparison is important in practice, since the strong performance of very large models must be balanced against issues of accessibility, efficiency and operational cost. We conduct experiments on two datasets containing questions of varying levels of difficulty, allowing us to evaluate the strengths and limitations of current LLMs in data-oriented analytical tasks.

The rest of the paper is organised as follows. Section 2 describes the evaluation methodology, Section 3 presents the experimental results, Section 4 discusses the core findings of the study, and finally, Section 5 concludes the paper.

## 2. Methodology

This section describes the methodology of the study. We present the datasets and questions used in the experiments, the examined tasks, the considered LLMs, the prompting strategies, and the evaluation metrics used to assess model performance.

### 2.1. Datasets

The first dataset corresponds to real survey data related to the acceptance of wind turbines in four countries. It contains answers to 16 questions related to wind turbine acceptance, as well as demographic information about the survey participants. The columns of the dataset are: COUNTRY, Q1, Q2, Q3, Q4_Env, Q5, Q6, Q7, Q8, Q9, Q10_Com, Q11, Q12, Q13, Q14, Q15_Ind, Q16_Global, AGE_GROUP, GENDER, EDUCATION, EMPLOYMENT, LIVED. The dataset contains 315 instances, each corresponding to a respondent. The data was collected in the context of the WIMBY project (HORIZON.2.5.2 - Energy Supply).

The second dataset contains sales data of a supermarket chain. The columns of the dataset are: sale_id, branch, city, customer_type, gender, product_name, product_category, unit_price, quantity, tax, total_price, reward_points. The dataset is publicly available on Kaggle (Name: SuperMarket Sales, Author: Chad Wambles).

### 2.2. Questions

Table 1 presents the 16 questions used for each of the datasets. The questions are grouped into four categories of increasing difficulty/complexity. Category Q1 corresponds to easy questions, while category Q4 corresponds to difficult questions requiring more complex data processing.

**Table 1. The questions used in the evaluation per dataset.**

| Questions for the SURVEY dataset | Questions for the SALES dataset |
|---|---|
| Q1.1 Which countries participated in the survey?<br>Q1.2 What are the first 10 entries in the survey?<br>Q1.3 How many people are employed full-time (EMPLOYMENT = "YES_FULL")?<br>Q1.4 How many respondents gave the answer 5 to Q1?<br>Q2.1 What is the average Q1 score for each country?<br>Q2.2 How many full-time employed people (EMPLOYMENT = "YES_FULL") are there in each country?<br>Q2.3 What are the maximum and minimum Q5 scores for each country? | Q1.1 How many transactions were made in total?<br>Q1.2 What is the average unit price of all products (column unit_price)?<br>Q1.3 How many transactions took place in Chicago?<br>Q1.4 What is the total sales amount (total_price) for the product “Shampoo”?<br>Q2.1 What are the total sales (total_price) for each city?<br>Q2.2 How many transactions were made for each product category (product_category)?<br>Q2.3 What are the total sales for each product category, sorted from highest to lowest?<br>Q2.4 Which gender has the highest average purchase quantity (quantity)? |
| Q2.4 Which education levels (EDUCATION) have the most respondents? | Q3.1 What percentage of customers are Members in each city? |
| Q3.1 Which countries have an average Q1 score greater than 3? | Q3.2 How many “Big Spender” transactions (total_price > 100) and “Regular Spender” transactions (total_price ≤ 100) were made in each city? |
| Q3.2 Which countries have more than 50 records? | Q3.3 What is the percentage of male and female customers in each city? |
| Q3.3 Which countries have an average Q3 score between 3 and 4? | Q3.4 Which branch-product category combinations have total sales above 5,000 dollars? |
| Q3.4 Which genders (GENDER) have an average Q5 score different from 3? | Q4.1 What is the average total price for Members only, for each city–product category–gender combination, considering only combinations with an average above 50 dollars? |
| Q4.1 Which countries have more records than the average number of records per country? | Q4.2 Which product has the highest total sales (Top 1) across all transactions? |
| Q4.2 Which countries have respondents with Q1 > 3 AND Q2 > 2 AND full-time employment (EMPLOYMENT = "YES_FULL")? | Q4.3 What is the average total_price for each combination of city, gender, and product category? |
| Q4.3 What is the average Q3 score for each combination of country, gender, and education level? | Q4.4 Which city has the highest average number of reward points (reward_points)? |
| Q4.4 Which country has the highest average Q16_Global score? | |

### 2.3. Tasks

We evaluate two different tasks: 1) *direct question answering*, where the model is asked to directly provide the answer to each question given the dataset, and 2) *SQL query generation*, where the model is asked to generate an SQL query corresponding to each question. In the second task, the generated query should produce the correct answer when executed over a specific database containing the data. Microsoft Access was used as the database system.

The first task allows us to examine the ability of LLMs to directly analyze a dataset and produce the correct answer. The second task evaluates the ability of LLMs to generate well-structured queries in SQL, a widely used language for querying relational databases.

### 2.4. Examined LLMs

We evaluate the performance of four large (state-of-the-art) LLMs: Claude Sonnet 4.6, ChatGPT 5.2, Grok 4.1, and Gemini 3. In addition, we include two smaller language models, Phi-3 Mini and Mistral 7B, which require fewer computational resources.

To prevent the models from retaining memory from previous interactions, we evaluated each prompt in a new session. For the large models, a new chat session was started for each configuration. For the smaller models, the session was terminated and the model was restarted, ensuring that the context window was cleared. This guaranteed that all experiments were conducted independently.

### 2.5. Prompts

We evaluate model performance using two different prompting strategies. The first strategy (*simple prompt*) provides limited information and context to the model. The second strategy (*detailed prompt*) provides more detailed instructions, sample data, and examples, aiming to guide the model more explicitly in performing the task.

### 2.6. Evaluation Metrics

We evaluate the accuracy of each combination of model and prompting strategy by computing the percentage of correct answers produced by the model. In addition, we evaluate efficiency by measuring the average time (in minutes) required by each model to produce the answer.

## 3. Results

In this section, we present the accuracy results for the two examined tasks (direct question answering and SQL query generation). The results are reported for each dataset, model, and prompting strategy, together with the average response time.

### 3.1 Direct Question Answering

Table 2 shows the accuracy of the models when answering questions directly over the datasets. The results clearly indicate the strong performance of the large models. ChatGPT 5.2 and Grok 4.1 achieved perfect accuracy (100%) across both datasets and prompting strategies. Gemini 3 and Claude Sonnet 4.6 also achieved very high accuracy, with average scores of 98.43% and 96.85%, respectively. The few observed errors were mainly related to rounding issues or minor differences in the provided answers (giving IDs instead of full entries). In contrast, the smaller models performed very poorly in this task. Phi-3 Mini produced no correct answers (0%), while Mistral 7B achieved only 1.56% accuracy, correctly answering a single question under the detailed prompt in the sales dataset. Response times also differed substantially across model categories. The large models typically produced answers within one to two minutes, while the smaller models required considerably longer processing times, often exceeding thirty minutes.

*Table 2. Accuracy per model and prompting strategy of the direct question answering task.*

| Model | SALES Simple | SALES Detailed | SURVEY Simple | SURVEY Detailed | Average accuracy | Average response time |
|---|---|---|---|---|---|---|
| **Claude Sonnet 4.6** | 16/16 (100%) | 16/16 (100%) | 15/16 (93.7%) | 15/16 (93.7%) | 96.85% | *~1 min* |
| **ChatGPT 5.2** | 16/16 (100%) | 16/16 (100%) | 16/16 (100%) | 16/16 (100%) | 100% | *~2 min* |
| **Grok 4.1** | 16/16 (100%) | 16/16 (100%) | 16/16 (100%) | 16/16 (100%) | 100% | *~2 min* |
| **Gemini 3** | 16/16 (100%) | 16/16 (100%) | 16/16 (100%) | 15/16 (93.7%) | 98.43% | *~1 min* |
| **Phi-3 Mini** | 0/16 (0%) | 0/16 (0%) | 0/16 (0%) | 0/16 (0%) | 0% | *~35 min* |
| **Mistral 7B** | 0/16 (0%) | 1/16 (6.25%) | 0/16 (0%) | 0/16 (0%) | 1.56% | *~30 min* |

An analysis of the responses showed that the high performance of the large models is related to their ability to execute code internally when processing datasets. During the experiments, the interfaces of several models indicated intermediate steps such as *"running code"*, revealing the use of an internal code execution environment. In practice, the models appear to read the dataset, generate Python code to process it, execute the code, and then return the results in natural language. This capability allows them to produce accurate answers without relying on error-prone textual reasoning. The smaller models lack such functionality. In many cases, instead of producing answers (as requested in the instructions), they generated Python code snippets intended to compute the results. In other cases, the models produced incorrect numerical values or hallucinated results. These observations explain the near-complete failure of the smaller models in the direct question answering task.

### 3.2 SQL Query Generation

Table 3 presents the results for the SQL query generation task. Overall, the large models again achieved exceptional accuracy. Claude Sonnet 4.6, ChatGPT 5.2, and Gemini 3 achieved 98.43% average accuracy, while Grok 4.1 achieved 96.85%. Response times were approximately one minute per set of questions, indicating consistent performance across tasks.

*Table 3. Accuracy per model and prompting strategy of the SQL generation task.*

| Model | SALES Simple | SALES Detailed | SURVEY Simple | SURVEY Detailed | Average accuracy | Average response time |
|---|---|---|---|---|---|---|
| **Claude Sonnet 4.6** | 15/16 (93.7%) | 16/16 (100%) | 16/16 (100%) | 16/16 (100%) | 98.43% | *~1 min* |
| **ChatGPT 5.2** | 16/16 (100%) | 16/16 (100%) | 16/16 (100%) | 15/16 (93.7%) | 98.43% | *~1 min* |
| **Grok 4.1** | 16/16 (100%) | 16/16 (100%) | 15/16 (93.7%) | 15/16 (93.7%) | 96.85% | *~1 min* |
| **Gemini 3** | 16/16 (100%) | 16/16 (100%) | 16/16 (100%) | 15/16 (93.7%) | 98.43% | *~1 min* |
| **Phi-3 Mini** | 10/16 (62.5%) | 10/16 (62.5%) | 12/16 (75%) | 8/16 (50%) | 62.5% | *~30 min* |
| **Mistral 7B** | 8/16 (50%) | 11/16 (68.75%) | 10/16 (62.5%) | 8/16 (50%) | 57.81% | *~25 min* |

The errors observed for the large models were rare and mostly related to SQL dialect incompatibilities and not incorrect logic. In some cases, the generated queries used SQL constructs that are supported in systems such as PostgreSQL but not in Microsoft Access, which was considered in our experiments (and mentioned in the prompts). Examples include the use of OVER, PARTITION BY, and TOP 1 WITH TIES. In these cases, the queries were logically correct but could not be executed. A small number of errors were also caused by typos. The smaller models performed considerably better in the SQL generation task than in the direct question answering task, but their accuracy remained significantly lower than that of the larger models. Phi-3 Mini achieved an average accuracy of 62.5%, while Mistral 7B achieved 57.81%. Response times were also much longer (25-30 minutes).

The errors of the smaller models fell into several categories. A common issue was the use of SQL syntax that is incompatible with Microsoft Access. Other errors involved incorrect placement of aggregate functions (e.g., using WHERE instead of HAVING), incomplete queries, or logical mistakes in the formulation of aggregation expressions. In some cases, the queries were syntactically correct but did not correspond to the intended logic of the question. We also observed that the performance of the smaller models varied depending on the complexity of the questions. Accuracy was relatively high for simpler questions but dropped significantly for more complex questions involving aggregation or conditional logic.

## 4. Discussion

The experimental results on the two tasks indicated important differences in the capabilities of the examined models. The large models achieved very high accuracy in both tasks, demonstrating their ability to analyze datasets and translate natural language questions into structured queries. The smaller models, however, performed very poorly in the direct question answering task but showed moderate performance in SQL query generation.

This difference can be explained by the nature of the two tasks and by the internal capabilities of the models. Direct question answering requires the model to perform numerical computations over the dataset. The strong performance of the large models appears to be closely related to their ability to internally generate and execute code for processing the data, a functionality that was not available in the smaller models. In contrast, SQL query generation mainly involves translating natural language questions into SQL syntax, relying more on prior training on large collections of SQL queries. We also observed that, in several cases, the smaller models produced hallucinated outputs, including invented numerical values or references to database tables and columns that were not present in the provided datasets. These responses were often presented with high confidence, which poses risks in practical data analysis scenarios.

Finally, response times also differed significantly across model categories. The large models typically produced answers within one to two minutes, while the smaller (locally deployed) models required substantially longer processing times, often exceeding twenty minutes.

This study also has certain limitations. The evaluation was conducted on only two datasets and a limited set of analytical tasks, which may not fully represent the diversity and complexity of real-world data analysis scenarios. In addition, some of the examined models are proprietary systems that evolve over time and may include undocumented internal functionalities, making exact reproducibility difficult. Therefore, the findings should be interpreted within the specific experimental settings considered, while future work could examine a broader range of datasets, tasks, and model configurations.

## 5. Conclusion

We evaluated the ability of LLMs to answer questions over datasets under two settings: direct question answering and SQL query generation. The results show that state-of-the-art LLMs achieve very high accuracy in both tasks, demonstrating strong capabilities in interpreting datasets and translating natural language questions into structured queries. In the direct question answering task, this performance appears to be supported by the models' ability to generate and execute code internally to process the data. In contrast, smaller models show significant limitations, particularly in tasks requiring direct numerical computation over datasets. These findings highlight both the potential of modern LLMs for supporting data analysis tasks and the current challenges faced by smaller (but more resource-efficient) models.